\documentclass[preprint,12pt,authoryear]{elsarticle}

\usepackage{amssymb}

\usepackage{graphicx,xcolor}
\usepackage{placeins}

\usepackage{subcaption}
\usepackage{tikz}
\usetikzlibrary{positioning, shapes, shadows, arrows}
\usetikzlibrary{decorations.pathreplacing}
\tikzstyle{abstract}=[circle, draw=black, fill=white]
\tikzstyle{labelnode}=[circle, draw=white,opacity=.2,text opacity=1]
\tikzstyle{invisiblenode}=[circle,dashed, inner sep=1pt,circle split,line width=1mm,minimum size=1.5cm]
\tikzstyle{line} = [draw, -latex']

\usepackage[normalem]{ulem} 

\newcounter{IEEE_VER}
\begin{document}

\begin{frontmatter}



\title{Adaptive Filters in Graph \\Convolutional Neural Networks}


\author{Andrea~Apicella}
\author{Francesco~Isgrò}
\author{Andrea~Pollastro}
\author{Roberto~Prevete}
\address{Laboratory of Artificial Intelligence, Privacy \& Applications (AIPA Lab)}
\address{Laboratory of Augmented Reality for Health Monitoring (ARHeMLab)}
\address{Department of Electrical Engineering and Information Technology\\University of Naples Federico II}

\begin{abstract}
\let\thefootnote\relax\footnotetext{This paper has been published in its final version on \textit{Pattern Recognition} journal with DOI \url{https://doi.org/10.1016/j.patcog.2023.109867} in Open Access mode. Please consider it as final and peer-reviewed version.}
Over the last few years, we have witnessed the availability of an increasing data generated from non-Euclidean domains, which are usually represented as graphs with complex relationships, and Graph Neural Networks (GNN) have gained a high interest because of their potential in processing graph-structured data. 
In particular, there is a strong interest in exploring the possibilities in performing convolution on graphs using an extension of the GNN architecture, generally referred to as Graph Convolutional Neural Networks (ConvGNN). 
Convolution on graphs has been achieved mainly in two forms: spectral and spatial convolutions. 
Due to the higher flexibility in exploring and exploiting the graph structure of data, there is recently an increasing interest in investigating the possibilities that the spatial approach can offer. 
The idea of finding a way to adapt the network behaviour to the inputs they process to maximize the total performances has aroused much interest in the neural networks literature over the years. 
This paper presents a novel method to adapt the behaviour of a ConvGNN to the input proposing a method to perform spatial convolution on graphs using input-specific filters, which are dynamically generated from nodes feature vectors. 
The experimental assessment confirms the capabilities of the proposed approach, which achieves satisfying results using a low number of filters. 
\end{abstract}



\begin{keyword}
Graph Convolutional Neural Networks \sep Deep Learning \sep Dynamic Neural Networks \sep Programmable ANNs \sep Graph Structure Learning
\end{keyword}

\end{frontmatter}


\section{Introduction}
\label{sec:introduction}
In the last few decades, Convolutional Neural Networks (CNNs) have gained much interest due to their potential and versatility in addressing a large scale of machine learning problems \citep{lecun1998gradient,lecun2015deep,hesser2021identification}, such as image processing \citep{krizhevsky2012imagenet,yu2021convolutional,yang2021accurate} and pattern recognition \citep{kim2014convolutional,cui2021geometric,lyu2021weakly} 
, while achieving great success. 
The potential of CNNs lies in extracting and processing local information performing convolution on input data using sets of trainable filters with a fixed size. 
However, the design of the convolution operation in the CNNs allows to process only regular data while, in the real world, there is a considerable amount of data that naturally lie on non-Euclidean domains, needing different techniques to be processed.

These data are often represented by \textit{graph-based} structures. 
Graph structures imply several difficulties in using standard data processing techniques, such as the impossibility of using classic CNNs due to the variable number of neighbours for each node (differently from regular data where the filter properties fix the number of neighbours for each node). 
This aspect has led to new processing techniques such as Graph Neural Networks (GNNs), which gained high interest during the last years.

First attempts \citep{sperduti1997supervised, gori2005new, scarselli2008_graph} of neural networks based on input graphs, generally referred to as Recurrent Graph Neural Networks (RecGNNs), were based on \textit{message passing} architectures, where an iterative process 
allows to learn, for each node, a representation of the relative neighbourhood information. 
Therefore, the learned node representations are used for classification or regression tasks.
However, as the size of the graphs increased, these approaches were more and more computationally expensive, and this represented a new challenge to overcome.

Due to the great success of CNNs, GNNs inherits convolution operation producing the Graph Convolutional Neural Networks (ConvGNNs), which have found their expression in two different approaches. 
The former are \textit{spectral} methods (see, for example, \citep{li2018_adaptive,levie2018_cayleynets}), that perform convolution based on graphs signal processing techniques. 
The latter are \textit{spatial} methods (see, for example, \citep{atwood2016_diffusion,hechtlinger2017_generalization}), that instead perform convolution using spatial information of data, similarly to what classical CNNs do.
ConvGNNs share the same idea of message passing with RecGNNs but in a non-iterative manner.

However, ConvGNNs are usually based on learned filters having constant values for each input fed to the network as well as classical convolutional networks.
However, ConvGNNs are usually based on learned filters having constant values for each input fed to the network as well as classical convolutional networks. 
In other words, the filter values are independent of the input values.
However, we note that adapting the Artificial Neural Network (ANN) inner behaviour in function of the input is an open research area in the scientific community. 
In a nutshell, while in standard approaches, the ANN input-output relationship, i.e., the ANN behaviour, after the training phase, is completely defined by a set of fixed network parameters (weights and biases). 
By contrast, the core idea of this research area is that the ANN behaviour also depends on the input itself or additional inputs.
A way to achieve this goal is setting the network parameters by another neural network which receives the same input of the former neural network or external/additional inputs (see, for example, \citep{donnarumma2012programming} and the hypernetworks proposed in \citep{ha2017hypernetworks,sun2017addressing}). 
Thus, ANN is able to dynamically change its behaviour according with the received inputs. 
In this paper, we refer to this type of approaches as \textit{Dynamic Behaviour Neural Networks} (DBNN).


In the last years, several works were proposed following DBNN approach \citep{ha2017hypernetworks,sun2017addressing,von2019continual}.
However, in the GNN field, to the best of our knowledge, the DBNN approach has not received too much attention. 
In significant research work  \citep{simonovsky2017_dynamic} the authors proposed the Edge-Conditioned Convolution (ECC) network, which performs spatial convolutions over graph neighbourhoods exploiting edge labels and generating input-specific filters from them.

In \citep{nachmani2019hyper}, the authors proposed a method to make the GNN message passing architecture adaptive to input nodes using an external hypernetwork.
Zhang et al. in \citep{zhang2018graph} proposed the Graph HyperNetwork (GHN), a model having weights generated by an hypernetwork exploiting a computation graph representation.
In \citep{balavzevic2019hypernetwork}, an hypernetwork architecture is proposed to generate relation-specific convolutional filters for convolution on graphs. 

In this paper, we exploit the possibility of dynamically changing the convolutional filter behaviour as a function of the input and propose a novel method to perform spatial convolution on graph-structured data. We will name our approach Dynamic Graph Convolutional Filters (DGCF). 
Following \citep{jia2016dynamic, simonovsky2017_dynamic}, convolutional filters will be generated using an external module, the \textit{filter-generating network}, that, during the training stage, learns to produce input-specific filters in order to perform an \textit{ad-hoc} filtering operation for each input sample.
Differently from \citep{simonovsky2017_dynamic}, in our approach filters are generated exploiting nodes feature vectors instead of edge labels.
Our approach is validated in three series of experiments, making a comparison with the standard convolution over $10$ repetitions, with randomly initialized weights for each repetition. 
Hypothesis tests are reported for each series of experiment to verify the significance of the results.
\newline
The advantages of the proposed approach can be summarised as follows:
\begin{itemize}
    \item It inherits the standard convolution operation from classical CNNs. The convolution is performed on each node with its nearest neighbours applying sets of fixed-sized filters.
    \item The network inner behaviour changes according to the input. Filters used for the convolution are dynamically generated using an external module based on the input graphs.
    \item Dynamic behaviour of the convolutional filters can lead to design simpler architectures than non-dynamic approaches with respect the number of convolutional filters, while leaving unaltered performances. Promising results can be achieved using a fewer number of convolutional filters than a non-dynamic approach.
    \item Training convergence can be reached in a fewer number of epochs. Using a dynamic approach, we empirically show that the learning stage needs a fewer number of epochs than using the non-dynamic approach.
\end{itemize}
This paper is organized as follows.
Section \ref{sec:related} briefly reviews the related literature; Section \ref{sec:method} describes the proposed method; the experimental assessment is described in Section \ref{sec:experim} while in Section \ref{sec:results} the obtained results are presented and discussed. In Section \ref{sec:analysis} a visual analysis of the training stage of our proposal is shown. The concluding Section \ref{sec:conclusion} is left to final remarks.
\section{Related works}
\label{sec:related}

In this section, we first report the related works in the context of DBNN approach and, then, we give a general description of the Graph Neural Networks, focusing on ConvGNNs.

\subsection{DBNN approaches}

The idea of controlling the behaviours of an ANN through the input itself or an additional/external input has a long history in the literature \citep{jordan1990attractor,schmidhuber1992learning,nishimoto2008learning,paine2004motor,noelle1995towards,siegelmann2012neural,eliasmith2005unified,bishop1994mixture,donnarumma2012programming}. 
For example, in \citep{paine2004motor} a set of external neural units, called \textit{control neurons}, are bidirectionally connected to all the neurons belonging to a lower layer network modulating their functions and favouring the generation of particular motor primitives. 
In \citep{bishop1994mixture,bishop1995neural}, the authors describe a way to represent general conditional probability densities 
by considering a parametric model for the distribution 
expressed as a neural network whose parameters are determined by the outputs of another neural network 
having the same input.
In \citep{ha2017hypernetworks} the filters of CNNs and LSTMs networks are generated by an auxiliary network. 

In \citep{jia2016dynamic} the authors defined the dynamic changes in ANNs' behaviours in the context of traditional CNNs using a proposed {\itshape dynamic filter module} to execute the convolution operation.
Dynamic filter module consists of two parts: a {\itshape filter-generating network}, that generates filters' parameters from a given input, and a {\itshape dynamic filtering layer}, that applies those generated filters to another input. 

In particular, the dynamic filtering layer can be instantiated as a {\itshape dynamic convolutional layer}, wherever the filtering operation is translation invariant. In \citep{jia2016dynamic}, 
considering two input images $I_A$ and $I_B$, not necessary different, the filter-generating network takes as input $I_A$ and outputs filters $\mathcal{F}_\theta$ to apply on $I_B$.
Filters $\mathcal{F}_\theta$ 
are parameterized by parameters $\theta$. 
In this way, an output $G=\mathcal{F}_\theta(I_B)$ is generated.
However, this method is developed in the context of classic CNNs; by contrast, in \citep{simonovsky2017_dynamic}, the authors attempt to perform a dynamic spatial convolution on graphs. 
The authors proposed the Edge-Conditioned Convolution (ECC), which uses a filter-generating network to output edge-specific filters for each input sample dynamically. 

The DGCF approach propsed in this paper is inspired by the work in  \citep{jia2016dynamic,simonovsky2017_dynamic}. 
We perform a convolution on input graphs using filters that are dynamically generated by a filter-generating network, thus obtaining a dynamical change in the behaviour of the ConvGNN. 
We point out that, differently from the ECC proposed in \citep{simonovsky2017_dynamic} where convolutional filters are \textit{edge-based}, our strategy (see Section \ref{sec:method}) considers \textit{node-based} filters, tweaking in this way the filtering operation on nodes by nodes themselves. 

Thus, summarizing, the proposed approach is different from ECC with respect to the input fed to the filter-generating network, and it differs from the other DBNN approaches since it is applied on graphs.



\subsection{Graph Neural Networks}
GNNs are showing positive effects in several applications, as for example road speed detection \citep{lu2020lstm}, molecular generation for drug discovery \citep{bongini2021molecular}, Point of Interest recommendation \citep{zhang2021leveraging} and others. In \citep{yu2021resgnet} the authors showed how GNNs can be useful in the COVID19 diagnosis exploiting underlying relations in chest images.
A first proposal of a neural network model for graph-structured data was made in \citep{scarselli2008_graph}. This model builds on the idea that graph nodes represent "concepts" related to each other via edges.
Each node $n$ is represented by a feature vector $x_{n}$ and each edge $(i,j)$ is described by a feature vector $\mathbf{x}_{(i,j)}^e$.
This model leverages information exchange among nodes and their neighbours to update their features iteratively (\textit{message passing} mechanism). 
In the literature, iterative graph processing techniques based on neural networks with a message passing architecture are generally referred as RecGNNs.
To face the computational costs of these methods, several kinds of neural network models were proposed in the literature, often with the aim of generalising classical and established neural networks data processing to graph data, such as \citep{goodfellow2014generative,liu2021neighbor,bahdanau2014neural,velickovic2017graph}.
{In \citep{wu2020comprehensive} a comprehensive survey on the topic is proposed}.\\
A particular focus was given in the literature to perform the  convolution operation on graph-structured data. 
Graph Convolutional Neural Networks (ConvGNNs) share the idea of message passing adopted by RecGNNs but implement it in a non-iterative manner: information is exchanged between neighbours using different convolutional layers, each with different filters \citep{wu2020comprehensive}.
However, the non-Euclidean characteristics of graphs (e.g., their irregular structure) makes the convolution and filtering operations not easy to define as for those on images. 
For this reason, in the past decades, researchers have been working on how to conduct convolution operations on graphs using several approaches, that can be categorized in:
\begin{itemize}
    \item  \textit{spectral approaches}, that rely on the graph spectral theory, involving graph signal processing, such as graph filtering and graph wavelets (see, for example, \citep{henaff2015_deep,li2018_adaptive,levie2018_cayleynets});
    \item \textit{spatial approaches}, that leverage on structural information to perform convolution, such as aggregations of graph signals within the node neighbourhood (see, for example, \citep{atwood2016_diffusion,hechtlinger2017_generalization}).
\end{itemize}
Although spectral architectures have been explored successfully in several works such as  \citep{henaff2015_deep,li2018_adaptive,levie2018_cayleynets},  one of the main problems of ConvGNNs in the spectral domain is that the graph structure has to be set for all the inputs due to the use of the graph Laplacian in the training stage. However, spectral analysis is computationally expensive, limiting the concrete usage of this methods on huge graphs \citep{balcilar2020bridging}.
Although strategies to use ConvGNNs with different inputs graph structures were proposed \citep{li2018_adaptive}, this problem is generally not present in the spatial domain.

For these reasons, several spatial domain methods have been proposed in the literature. For example, in \citep{niepert2016_learning} the authors present PATCHY-SAN, a ConvGNN model inspired by the classical image-based CNN. In \citep{atwood2016_diffusion} and \citep{hechtlinger2017_generalization} two different methods to generalise the convolutional operator using random walks for neighbourhood locating were reported. In \citep{fu2019hplapgcn} spectral-based GNNs are generalized to work on data structured as hypergraphs instead of classical graphs.
In \citep{wu2019simplifying} the complexity of a GNN was reduced, collapsing the network layers into a single linear transformation.
In \citep{zhang2019_learning} the authors proposed a method to learn or refine the graph structure together with the network parameters.
\newline
\newline
The aim of the work presented in this paper is to perform an adaptive spatial-convolution, using fixed-sized filters, on graph structured data.
According to the dynamic convolutional layer proposed in \citep{jia2016dynamic}, in our approach, for each sample, a translation invariant set of filters is generated by a filter-generating network and shared among all the neighbourhoods. 
\section{Method description}
\label{sec:method}
In this work, we propose a ConvGNN-based architecture whose convolutional filters change in function of the input features. We name them Dynamic Graph Convolutional Filters (DGCF). Differently from similar works as \citep{simonovsky2017_dynamic}, where filters depend on the graph edges, we propose a DBNN approach
based on the graph nodes' features. 
In this section, after a brief introduction of graphs' notation, a detailed description of our proposal is given.

\subsection{Notations}
Let $G = (V, E)$ be an undirected or directed graph where $V$ is a finite set of $N$ nodes, and $E$ is a finite set of edges. 
We define in boldface $\mathbf{x^{i}} \in \mathbb{R}^{1 \times J}$ the input feature vector related to the node $i \in V$, where $J$ is the number of input channels, and $\mathbf{y^{i}} \in \mathbb{R}^{1 \times M}$ its output feature vector, where $M$ is the number of output channels. 
Let $X \in \mathbb{R}^{N \times J}$ denote the matrix representation of an input graph as an embedding of the feature vectors of its nodes.

In order to obtain neighbourhoods with a sufficient number of nodes to which apply a filter of dimension $K$, we select the \textit{k-nearest neighbours} of each node using the classical shortest path distance \citep{buckley1990distance}. Neighbourhoods are uniquely defined for each node. 

\subsection{Dynamic Graph Convolutional Filters}
This work aims to perform convolution on graphs using dynamically generated filters conditioned on a given input. As we have described above, otherwise from ECC in \citep{simonovsky2017_dynamic}, where convolution is performed using dynamical \textit{edge-based} filters, our intent consists in using \textit{node-based} filters dynamically generated from nodes' feature vectors.

Considering the matrix representation $X$ of an input graph, 
using a neural network $h_\theta(\cdot)$, that we will refer as filter-generating network, 
with parameters $\theta$, 
we can generate a set of node-specific filters $\mathcal{F}=h_{\theta}(X)$ used to compute a dynamic convolution on input graphs.
$\mathcal{F}$ can be represented as a matrix $\mathcal{F} \in \mathbb{R}^{J \times K \times M}$, where $J$ is the number of input channels, $K$ is the filter size and $M$ is the number of output channels. Supposing to compute the $m$-th output channel of the node $n$, what we propose can be formalised as follows:
$$\mathbf{y}^{n}_{m}=f\Big(\sum\limits_{j=1}^{J}\sum_{k=1}^{K}\limits \mathcal{F}_{jkm} \mathbf{x}_{j}^{s(n,k)} \Big)$$ 
where $s(n,k)$ returns the index of the $k$-th neighbour of $n$, $F_{jkm}$ are the filters generated by the  filter-generating network $h_{\theta}(\cdot)$. In other words, the $F_{jkm}$ is the value in position $(j,k)$ 
of the $m$-th filter generated by the filter-generating network $h_\theta(\cdot)$. 
Thus, during the training stage, the parameters $\theta$ have to be learned, together with the other network's parameters. 
According to what is described in \citep{jia2016dynamic}, our approach follows the \textit{dynamic convolutional layer}: the filter-generating network, defined as $h_{\theta}:\mathbb{R}^{N \times J} \longrightarrow \mathbb{R}^{J \times K \times M}$, where $N$ is the number vertices, $J$ is the number of input channels, $K$ is the filter size and $M$ is the number of the output channels,
takes as input the the input graph and generates a unique set of filters shared among all the neighbourhoods.
As we said above, we will refer to our proposal as Dynamic Graph Convolutional Filters (DGCF).

\begin{figure*}[!ht]
	\centering
	\includegraphics[width=0.9\textwidth]{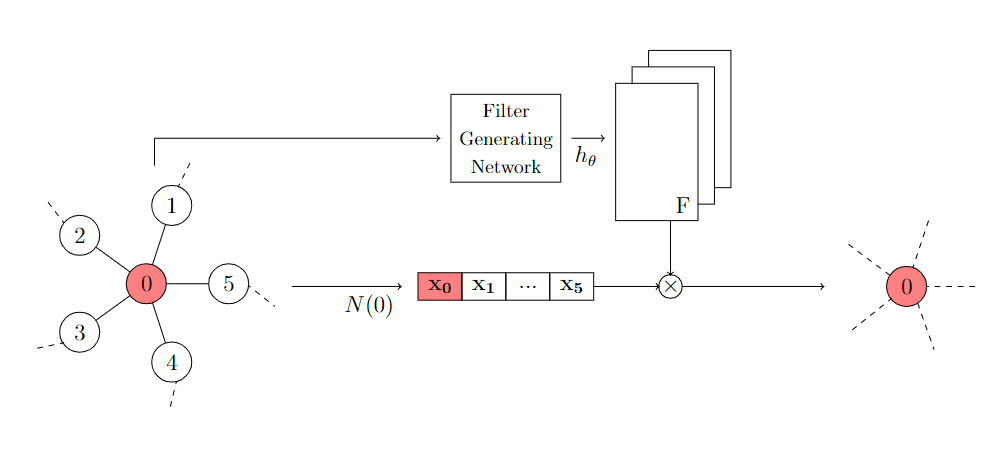}
    \caption{A graphical description of the DGCF layer in processing a node - marked in red and labeled as $0$ - using filters of dimension $K=5$.
    A unique set of filters dependent on all the nodes' feature vectors of the input graph is dynamically generated and shared among all of its neighbourhoods. 
    Finally, input feature vectors of the neighbourhoods are weighted summed to compute the nodes' outputs.
    In this example, input-specific filters are applied on the neighbourhood of the node 0, referred as $N(0)$.
    }
    \label{figure:pipeline}
\end{figure*}

A graphical representation of the DGCF layer is shown in Figure \ref{figure:pipeline}: supposing to have a node, labelled as $0$, as a central node during the convolution operation on a given input graph, an input-specific set of filters is firstly generated by the filter-generating network using the nodes' feature vectors of the input graph, then it is applied to the neighborhood of the node $0$, referred to as $N(0)$, computing a new representation of it. This procedure is then iterated over all the nodes of the input graph.
From the experimental results, as reported in Section \ref{sec:results}, emerged that the use of few dynamic convolutional filters DGCF leads to results comparable with traditional convolutional architectures composed of an higher number of filters.
\newcommand{\stargraph}[2]{
\begin{tikzpicture}
    \node[draw,circle,fill=red!50] at (360:0mm) (center) {0};
    \foreach \n in {1,...,#1}{
        \node[draw,circle] at ({\n*360/#1}:#2cm) (n\n) {$\n$};
        \draw (center)--(n\n);
        \draw[dashed] ({\n*360/#1}:#2cm+9)--({\n*350/#1}:#2cm+25);
    }
\end{tikzpicture}
}
\newcommand{\stargraphcomplete}[2]{
\begin{tikzpicture}
    \node[draw,circle,fill=red!50] at (360:0mm) (center) {0};
    \foreach \n in {1,...,#1}{
        \node[draw,circle] at ({\n*360/#1}:#2cm) (n\n) {$\n$};
        \draw (center)--(n\n);
    }
    \foreach \n in {1,...,#1}{
        \draw (n\n)--({\n*350/#1}:#2cm+25);
        \node[draw,circle] at ({\n*350/#1}:#2cm+34){\the\numexpr\n+#1};
    }
\end{tikzpicture}
}

\section{Experimental assessment}
\label{sec:experim}
The DGCF approach was experimentally evaluated in three series of experiments. 
We point out that our interest in these experiments is investigating the advantages of using our method to non-dynamic approaches in terms of results, learning time, and model simplicity.

At first, we conducted preliminary experiments on the well-known benchmark dataset MNIST, 
representing images in terms of graphs (see, \citep{hechtlinger2017_generalization})
Then, we conducted a series of experiments on the 20NEWS \citep{joachims1996probabilistic}, a commonly used dataset in the ConvGNN literature.
Finally, we conducted a series of experiments on an widely used electroencephalographic (EEG) signals dataset, SEED \citep{zheng2015investigating}, in order to test our method on harder tasks, such as emotion recognition from EEG signals. 

On this last task, a model analysis was made in order to investigate the functioning of the DGCF layer. In particular, a comparison between the learning processes (in terms of weights updates for each epoch) between the static filters and the filters generated by the filter-generating network was carried out.
Note that, in each series of experiments we selected a non-dynamic ConvGNN reported in the literature, trained on the same dataset we used. 
For each series of experiments we selected a reference paper using the same dataset from the literature on non-dynamic ConvGNN, and used the same topology and experimental setup reported in the paper.

\subsection{MNIST}
We used the widely known MNIST handwritten digits dataset for running the first series of experiments.
The MNIST dataset consists in $70000$ grayscale images of handwritten digits. This dataset was reported already split in training ($60000$ samples) and test ($10000$ samples) sets.
So, recognizing each digit can be viewed as a 10-classes classification problem.
We adopted the same experimental setup on the MNIST dataset used in \citep{hechtlinger2017_generalization} that can be resumed as follows: after the exclusion of constant pixels, the correlation matrix $C$ between pixels is computed to estimate their relationships. Therefore, two pixels (nodes) $i,j$  are considered connected if $|C_{ij}|>T$ where $T$ is a fixed threshold.

\subsection{20NEWS dataset}
The 20NEWS dataset consists of $17236$ text documents, labeled using $20$ classes. The original  split in training  set ($10167$ samples after the preprocessing) and test set ($7069$ samples after the preprocessing) was adopted in this work. This task can be defined as a 20-classes classification problem.
Following the preprocessing described in \citep{defferrard2016_convolutional,zhang2019_learning}, we considered only the $10000$ most frequently used words, considering the bag-of-words model to represent each document. We used \textit{word2vec} \citep{mikolov2013efficient} to represent each word as a vector. 
Finally, the cosine similarity metric between words vectors was adopted to compute the connections between words.

\begin{table}[!h]
\centering
\begin{tabular}{c|c|c|c}
	\hline
	Dataset  & $\#graphs$ & $\#nodes$ & $\#classes$\\
    \hline \hline
    MNIST & 70000 & 717 & 10\\
    20NEWS & 17236 & 10000 & 20\\
    SEED & 50910 & 62 & 3\\
    \hline
\end{tabular}
\caption{Datasets selected for the experiments.}
\label{table:datasets}
\end{table}

\subsection{SEED}
The SEED dataset \citep{zheng2015investigating} consists of EEG signals recorded from 15 subjects while they were watching video clips of about 4 minutes. Each video clip was carefully chosen to induce three types of emotions, i.e. negative, neutral and positive. For each subject, three sessions of 15 trials/video clips were collected. EEG signals were recorded in 62 channels using the ESI Neuroscan System\footnote{https://compumedicsneuroscan.com/}. As in \citep{zhong2020eeg}, we consider the pre-computed differential entropy (DE) features smoothed by linear dynamic systems (LDS). DE features are pre-computed, for each second, in each channel, over the following five bands: delta (1–3 Hz); theta (4–7 Hz); alpha (8–13 Hz); beta (14–30 Hz); gamma (31–50 Hz). 

Samples were modeled as graphs considering each EEG channel as a node, and the DE features related to the 5 bands as feature vector 
of each node.
The adjacency matrix $A \in \mathbb{R}^{n \times n}$ was modeled considering the EEG channels disposition on the scalp, where $n$ represents the number of channels in EEG signals. In particular, each entry $A_{ij}$ represents the physical distance between the sensor $i$ and the sensor $j$, computed referring to the International 10/20 Positioning System. According to this system, electrodes are placed at a distance of 10\% (or 20\%) of the Inion-Nasion distance \citep{myslobodsky1989locations}.
Table \ref{table:datasets} reports a summary of the datasets used in our experiments. 

\subsection{Experimental setup}
The proposal was validated by analyzing its impact on existing architectures in literature \citep{hechtlinger2017_generalization, zhong2020eeg, zhang2019_learning}. For each architecture, we evaluated the models as the number of convolutional filters changes, as it is shown in the relative configuration (Tables \ref{table:mnist_configs}, \ref{table:20news_configs}, \ref{table:seed_configs}).
For the experiments on the MNIST and 20NEWS datasets, model performance estimation was performed using an \textit{holdout method} since both of the datasets report a predefined train/test split. Moreover, model performances were evaluated based on the performances averaged over 10 repetitions, where, for each repetition, models' parameters were reinitialized following the same inizialization criterion.

For the experiments on the SEED dataset, model performance estimation was 
estimated
focusing on the \textit{Subject-Independent Classification}: still following the experimental protocol of \citep{zhong2020eeg}, we adopted the \textit{leave-one-subject-out cross-validation}, in which 14 subjects are considered as training set, and the remaining subject as test set. This was repeated for each possible configuration. The performance is evaluated averaging the test accuracies using one session of data. 
Moreover, for each experiment, during the training stage, 30\% of the training set was extracted following a \textit{stratified sampling} \citep{parsons2014stratified}.
Each experiment was run considering \textit{early stopping} as convergence criterion.
Significance differences about the comparisons between the models were tested using \textit{hypothesis testing}. For each result - expressed by the average and the standard deviation - a normality test was firstly made using the \textit{Shapiro-Wilk} test \citep{shaphiro1965analysis}. Then, according to the results of the normality tests, hypothesis tests were made using the Student's \textit{t-test} \citep{student1908probable}, in the case of normally distributed data, or Mann-Whitney \textit{U-test} \citep{mann1947test}, otherwise. For each test, a significance level of $\alpha = 0.05$ was considered.
Hypothesis tests were formulated as follows:
$$H_0: \mu_{ConvGNN} \ge \mu_{DGCF}$$
$$H_1: \mu_{ConvGNN} < \mu_{DGCF}$$
\textit{Full-connected neural networks} as filter-generating network (FGN) architectures were adopoted, whose number of nodes and layers was tuned using a \textit{bayesian optimization} \citep{snoek2012practical} in each experiment.
\section{Results}
\label{sec:results}

In each experiment, the adopted models were two layers architectures $L_1-L_2$, $L_i\in\{Ct, DCt, FCn, Pooling\}$, where $Ct$, $DCt$ referred to a $t$ filters static ConvGNN layer and a $t$ filters DGCF layer respectively, $FCn$ referred to a fully connected layer having $n$ hidden units, and $Pooling$ referred to a global pooling layer \citep{xu2018powerful}.

\subsection{MNIST}
Using the MNIST dataset, the configuration $C20-C20$ proposed by \citep{hechtlinger2017_generalization} is used. 
In the experimental setup of this work, a comparison was made substituting the first convolutional layer with a dynamic one.
In our analysis, we varied the number of convolutional filters of the first layer in order to evaluate the effectiveness in using a dynamical approach, in both the static and dynamic models. A summary of the adopted configurations is shown in Table \ref{table:mnist_configs}.

\begin{table}[!h]
\fontsize{9pt}{10.5pt}\selectfont
\centering
\begin{tabular}{c|c|c}
	\hline
	Hyperparameter & DGCF & ConvGNN\\
    \hline \hline
    Architecture & $DCi-C20, i \in \{2,4,8,16,20\}$ & $Ci-C20, i \in \{2,4,8,16,20\}$\\
    $K$ & 6 & 6\\
    Optimizer & Adam & Adam\\
    Learning Rate & $10^{-3}$ & $10^{-3}$\\
    FGN Architecture & $FC200$ & -\\
    \hline
\end{tabular}
\caption{Configurations used for the experiments on MNIST dataset.
We denote with $Ct$ a ConvGNN layer with $t$ filters, with $DCt$ a DGCF layer with $t$ filters, and with $FCn$ a fully connected layer with $n$ hidden units.
}
\label{table:mnist_configs}
\end{table}
\begin{figure}[!h]
     \centering
     \begin{subfigure}[b]{0.45\textwidth}
         \centering
         \includegraphics[width=\textwidth]{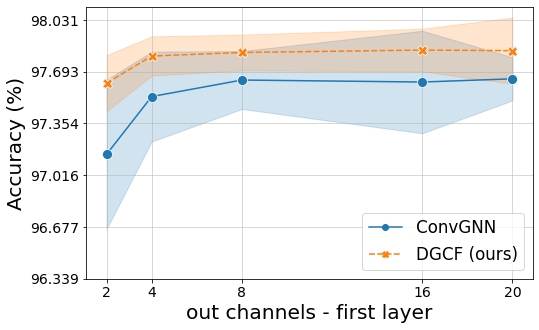}
         \caption{}
     \end{subfigure}
     \hfill
     \begin{subfigure}[b]{0.45\textwidth}
         \centering
         \includegraphics[width=\textwidth]{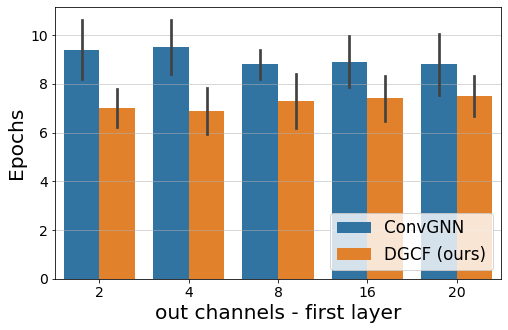}
         \caption{}
     \end{subfigure}
        \caption{Results of the experiment on MNIST dataset: (a) mean and standard deviation band of accuracy per configuration; (b) average of the training epochs to convergence per configuration.}
        \label{fig:mnist_results_images}
\end{figure}

As it is shown in Figure \ref{fig:mnist_results_images}(a), the introduction of the dynamical layer increases the average performance of the architecture. Moreover, its interesting to notice that our dynamic approach leads to good performances also with simpler architectures: results comparable with the ones reported in \citep{hechtlinger2017_generalization} are achieved using a fewer number of convolutional filters. It is also important to point out that using our approach, convergence during the training stage was reached in a fewer number of epochs, as it is shown in Figure \ref{fig:mnist_results_images}(b). In Table \ref{table:mnist_hypothesis_tests} the results related to the hypothesis tests are reported: the null hypothesis was rejected for each configuration ($p < 0.05$), confirming the significance of the improvement given by our approach.

\begin{table}[!h]
\centering
\begin{tabular}{cc}
	\hline
    \multicolumn{2}{c}{Hypothesis tests}\\
	\hline
	Configuration & $p$-value\\
    \hline \hline
    $C2-C20$ & $0.0152$ \\
    $C4-C20$ & $0.0146$ \\
    $C8-C20$ & $0.0066$ \\
    $C16-C20$ & $0.0322$ \\
    $C20-C20$ & $0.0420$ \\
    \hline
\end{tabular}
\caption{Hypothesis tests results for the MNIST dataset.}
\label{table:mnist_hypothesis_tests}
\end{table}

\subsection{20NEWS}
On these data, we referred to the architectures presented in \citep{zhang2019_learning}, made by $C16-FC100$. Also in this case, the comparisons were made varying the number of convolutional filters, and considering the convolutional layer both as static and dynamic. A summary of the adopted configurations is shown in Table \ref{table:20news_configs}.

\begin{table}[!h]
\fontsize{9pt}{10.5pt}\selectfont
\centering
\begin{tabular}{c|c|c}
	\hline
	Hyperparameter & DGCF & ConvGNN\\
    \hline \hline
    Architecture & $DCi-FC100, i \in \{2,4,8,16\}$ & $Ci-FC100, i \in \{2,4,8,16\}$\\
    $K$ & 25 & 25\\
    Optimizer & Adam & Adam\\
    Learning Rate & $10^{-3}$ & $10^{-3}$\\
    FGN Architecture & $FC100-FC100$ & -\\
    \hline
\end{tabular}
\caption{Configurations used for the experiments on 20NEWS dataset. We denote with $Ct$ a ConvGNN layer with $t$ filters, with $DCt$ a DGCF layer with $t$ filters, and with $FCn$ a fully connected layer with $n$ hidden units.}
\label{table:20news_configs}
\end{table}
\begin{figure}[!h]
     \centering
     \begin{subfigure}[b]{0.45\textwidth}
         \centering
         \includegraphics[width=\textwidth]{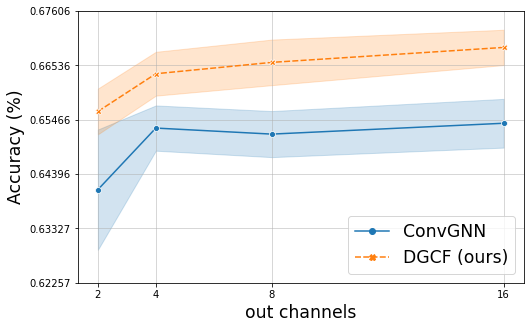}
         \caption{}
     \end{subfigure}
     \hfill
     \begin{subfigure}[b]{0.45\textwidth}
         \centering
         \includegraphics[width=\textwidth]{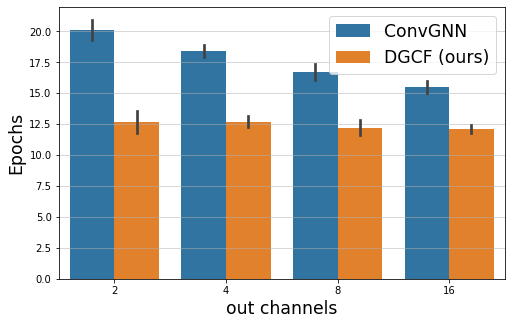}
         \caption{}
     \end{subfigure}
        \caption{Results of the experiment on 20NEWS dataset: (a) mean and standard deviation band of accuracy per configuration; (b) average of the training epochs to convergence per configuration.}
        \label{fig:20news_results_images}
\end{figure}

As it is shown in Figure \ref{fig:20news_results_images}(a), the introduction of the dynamical layer increases the average performances. In Figure \ref{fig:20news_results_images}(b) we can observe again how our method has a faster convergence than the static one.
In Table \ref{table:20news_hypothesis_tests} the results related to the hypothesis tests are reported: in each case, the null hypothesis was rejected confirming the significance of the improvement given by our method.

\begin{table}[!h]
\centering
\begin{tabular}{cc}
	\hline
    \multicolumn{2}{c}{Hypothesis tests}\\
	\hline
	Configuration & $p$-value\\
    \hline \hline
    $C2-C20$ & $0.0015$ \\
    $C4-C20$ & $0.0010$ \\
    $C8-C20$ & $0.0010$ \\
    $C16-C20$ & $< 0.0001$ \\
    \hline
\end{tabular}
\caption{Hypothesis tests results for the 20NEWS dataset.}
\label{table:20news_hypothesis_tests}
\end{table}

\subsection{SEED}
The base architecture considered for this experiment was similar to the RGNN model proposed by \citep{zhong2020eeg}, i.e. $Ct-Pooling$. Differently from \citep{zhong2020eeg}, in this series of experiment domain adaptation \citep{wang2018deep} techniques were not adopted. Moreover, we chose the global mean pooling \citep{xu2018powerful} across all the nodes on the graph instead of sum pooling, since it gave better results in firsts exploratory experiments. Also in this case, comparisons were made considering the convolutional layer both as static and dynamic, and varying the number of convolutional filters. Finally, the weight decay parameter was introduced to decrease the models' complexity.
A summary of the adopted configurations is shown in Table \ref{table:seed_configs}.

\begin{table}[!h]
\fontsize{9pt}{10.5pt}\selectfont
\centering
\begin{tabular}{c|c|c}
	\hline
	Hyperparameter & DGCF & ConvGNN\\
    \hline \hline
    Architecture & $DCi-AvgPool, i \in \{2,4,8,16,32\}$ & $Ci-AvgPool, i \in \{2,4,8,16,32\}$\\
    $K$ & 9 & 9\\
    Optimizer & Adam & Adam\\
    Learning Rate & $10^{-3}$ & $10^{-3}$\\
    FGN Architecture & $FC100-FC100-FC100$ & -\\
    \hline
\end{tabular}
\caption{Configurations used for the experiments on SEED dataset. We denote with $Ct$ a ConvGNN layer with $t$ filters, with $DCt$ a DGCF layer with $t$ filters, and with $FCn$ a fully connected layer with $n$ hidden units.}
\label{table:seed_configs}
\end{table}
\begin{figure}[!h]
     \centering
     \begin{subfigure}[b]{0.45\textwidth}
         \centering
         \includegraphics[width=\textwidth]{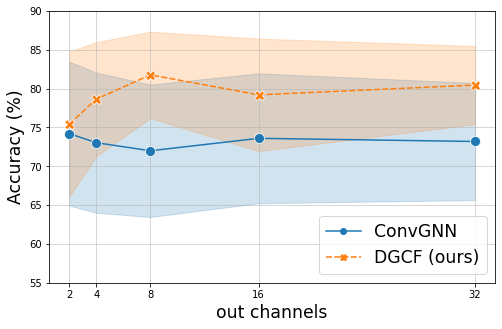}
         \caption{}
     \end{subfigure}
     \hfill
     \begin{subfigure}[b]{0.45\textwidth}
         \centering
         \includegraphics[width=\textwidth]{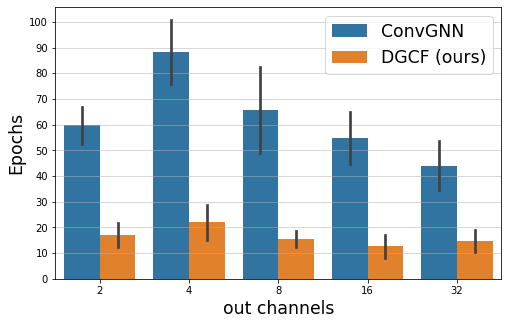}
         \caption{}
     \end{subfigure}
        \caption{Results of the experiment on SEED dataset: (a) mean and standard deviation band of accuracy per configuration; (b) average of the training epochs to convergence per configuration.}
        \label{fig:seed_results_images}
\end{figure}

As it is shown in Figure \ref{fig:seed_results_images}(a), the introduction of the dynamical layer strongly increases the average performance of the architecture. In Figure \ref{fig:seed_results_images}(b) it is enhanced the quicker convergence of our method than the static one. In Table \ref{table:seed_hypothesis_tests} the results related to the hypothesis tests are reported: except for the case $C2$, the null hypothesis was rejected for each configuration, confirming that the improvement given by our method is significant.
\begin{table}[!h]
\centering
\begin{tabular}{cc}
	\hline
    \multicolumn{2}{c}{Hypothesis tests}\\
	\hline
	Configuration & $p$-value\\
    \hline \hline
    $C2$ & $0.3118$ \\
    $C4$ & $0.0138$ \\
    $C8$ & $< 0.0001$ \\
    $C16$ & $0.0088$ \\
    $C32$ & $0.0007$ \\
    \hline
\end{tabular}
\caption{Hypothesis tests results for the SEED dataset.}
\label{table:seed_hypothesis_tests}
\end{table}
It's interesting to notice what is shown in Table \ref{table:seed_comparison_zhong}: referring to what is presented in \citep{zhong2020eeg}, our method overcomes some domain adaptation techniques, such as TCA.
\begin{table}[!h]
\centering
\begin{tabular}{cc}
	\hline
	\multicolumn{2}{c}{SEED}\\
	\hline
	Model & Accuracy (mean $\pm$ std)\\
    \hline \hline
    SVM & $56.73 \pm 16.29$ \\
    TCA* \citep{pan2010domain} & $63.64 \pm 14.88$ \\
    SA* \citep{fernando2013unsupervised} & $69.00 \pm 10.89$ \\
    T-SVM \citep{collobert2006large} & $72.53 \pm 14.00$ \\
    DGCNN \citep{song2018eeg} & $79.95 \pm 09.02$ \\
    DAN* \citep{li2018cross} & $83.81 \pm 08.56$\\
    BiDANN-S* \citep{li2018bi} & $84.14 \pm 06.87$\\
    BiHDM* \citep{li2020novel} (SOTA) & $85.40 \pm 07.53$\\
    RGNN* \citep{zhong2020eeg} & $85.30 \pm 06.72$\\
    \hline
    DGCF (ours) & $81.76 \pm 05.38$\\
    \hline
\end{tabular}
\caption{Subject-independent best average classification accuracy (mean $\pm$ std) on SEED dataset using different methods, as reported in \citep{zhong2020eeg}. In the last row, the best average accuracy of the proposed model was reported.
Methods highlighted with * involve the use of Domain Adaptation techniques.}
\label{table:seed_comparison_zhong}
\end{table}
\section{Model Analysis on SEED dataset}
\label{sec:analysis}
In this section we propose a visual analysis of the training stages of both the dynamic and static approach on the SEED dataset experiment.
The main aim of this analysis consists in inspecting differences in filter generation according to the input. In particular, assuming that common patterns are shared among equally-labeled samples, we expect that filters generated for samples of the same class are similar to each other, while they are different for samples belonging to different classes. 
We chose the SEED dataset for this analysis since it has the lowest number of classes among the datasets used for the experimental assessment.
This analysis was made considering a random subject as test set, and the remaining as training set. Configurations $C2-AvgPool$, for the static model, and $DC2-AvgPool$, for the dynamic model, were considered. The remaining details about the experimental setup follow the Table \ref{table:seed_configs}.

\begin{figure}[!h]
     \centering
     \begin{subfigure}[b]{0.3\textwidth}
         \centering
         \includegraphics[trim={0 0 0 2cm},clip,width=\textwidth]{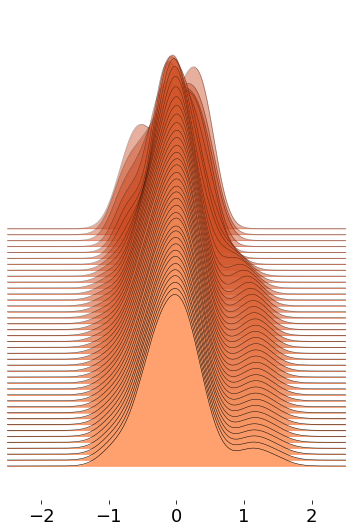}
         \caption{}
     \end{subfigure}
     \begin{subfigure}[b]{0.3\textwidth}
         \centering
         \includegraphics[trim={0 0 0 2cm},clip,width=\textwidth]{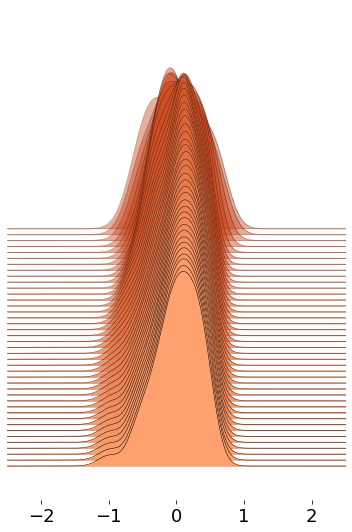}
         \caption{}
     \end{subfigure}
    \caption{Ridgeplot weights representation of a static convolutional layer with two filters during the learning process on the SEED dataset. Histograms related to the weights assumed by the first (a) and the second (b) filter are reported for each epoch.
    Histograms are overlapped from the first (background) to the last epoch (foreground).
    }
    \label{fig:seed_viz_filter_learning_static}
\end{figure}

\begin{figure}[!h]
    \centering
    \begin{subfigure}[b]{0.9\textwidth}
        \centering
        \includegraphics[width=0.31\linewidth]{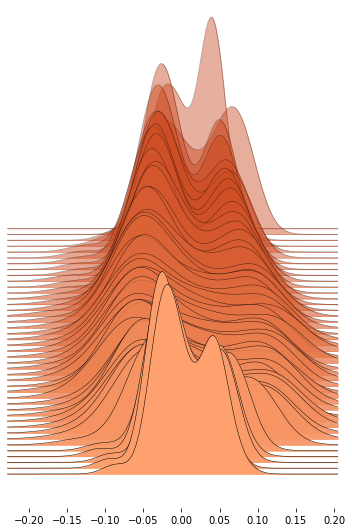}
        \hfill
        \includegraphics[width=0.31\linewidth]{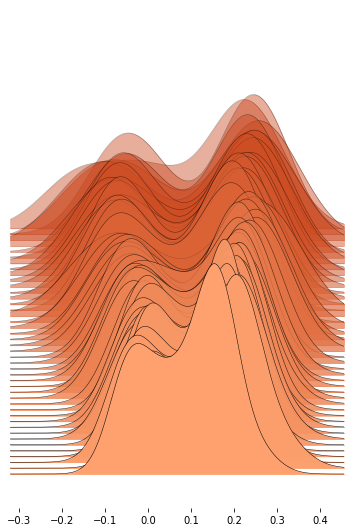}
        \hfill
        \includegraphics[width=0.31\linewidth]{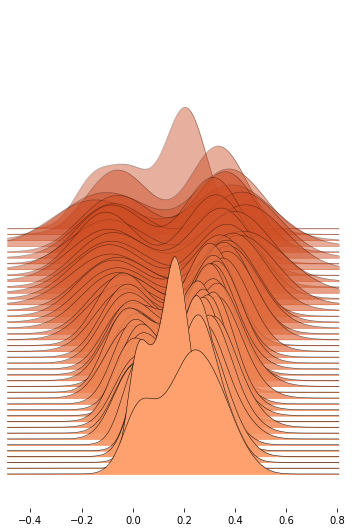}
        \caption{}
    \end{subfigure}
    \vskip\baselineskip
    \begin{subfigure}[b]{0.9\textwidth}
        \centering
        \includegraphics[trim={0 0 0 3cm},clip,width=0.31\linewidth]{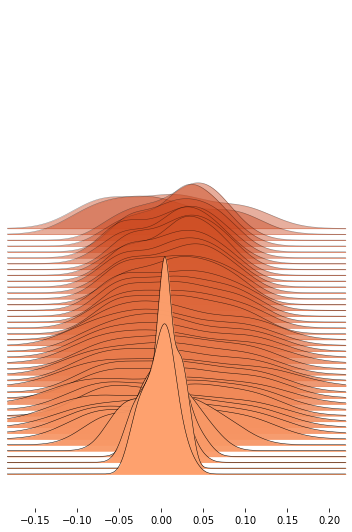}
        \hfill
        \includegraphics[trim={0 0 0 3cm},clip,width=0.31\linewidth]{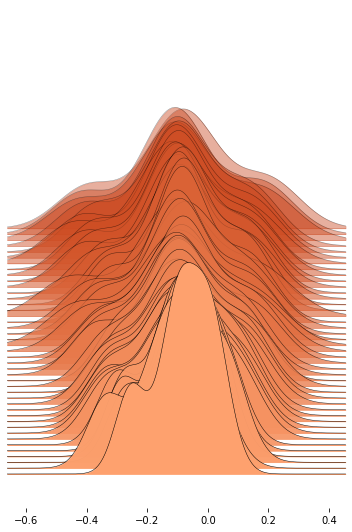}
        \hfill
        \includegraphics[trim={0 0 0 3cm},clip,width=0.31\linewidth]{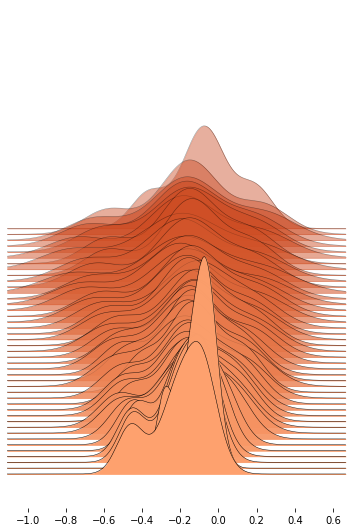}
        \caption{}
    \end{subfigure}
    \caption{Ridgeplot weights representation of a dynamic convolutional layer with two filters during the learning process on the SEED dataset. Histograms related to the weights assumed by the first (a) and the second (b) filter are reported for each epoch.
    Histograms are overlapped from the first (background) to the last epoch (foreground).
    For each filter, training process related to the negative (left), neutral (center) and positive (right) labels are represented. 
    The histogram in foreground is related to the final filter configuration.}
    \label{fig:seed_viz_filter_learning_dynamic}
\end{figure}

The learning processes are graphically described by weights distribution after each training epoch. In Figure \ref{fig:seed_viz_filter_learning_static}, the training of both the static filters are shown.
In Figure \ref{fig:seed_viz_filter_learning_dynamic}, the training stage of both of the dynamic filters, produced by the filter-generating network, are shown for each label (negative, neutral and positive, from left to right). Since the filters are generated uniquely for each sample, in order to have a fair evaluation of how they are distributed for each epoch, filters related to correctly classified samples 
in each epoch are collected and averaged.

From Figure \ref{fig:seed_viz_filter_learning_dynamic}, we can observe how the filter generation process is different for each class: assuming that equally-labeled data should share common patterns, Figure \ref{fig:seed_viz_filter_learning_dynamic} shows how different ranges of values are covered on the final parameters' configuration for input belonging to different classes, involving that specific 
filters are obtained according to its features. The direct consequence of this aspect is that feature extraction of the main architecture is enhanced by intrinsic patterns hidden in the sample itself. In facts, we can see that, almost in all cases, satisfying results are achieved using a low number of filters.
Moreover, comparing Figures \ref{fig:seed_viz_filter_learning_static} and \ref{fig:seed_viz_filter_learning_dynamic}, we can notice a difference in the ranges of values covered by the last configurations of both the models: static filters have weights included in the range $[-1, 1.5]$, while the dynamic ones have weights overall included in the range $[-0.6, 0.5]$. This aspect involves that the use of the dynamic layer could lead to less complex models, thus avoiding over-fitting/under-fitting \citep{bishop2014bishop}. 

\begin{figure}[!h]
     \centering
     \begin{subfigure}[b]{0.3\textwidth}
         \centering
         \includegraphics[width=\textwidth]{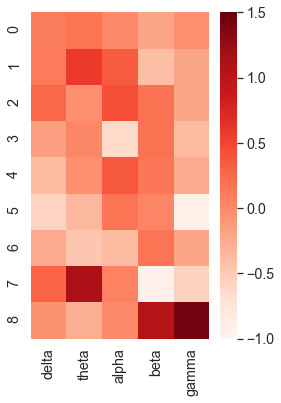}
         \caption{}
     \end{subfigure}
     \begin{subfigure}[b]{0.3\textwidth}
         \centering
         \includegraphics[width=\textwidth]{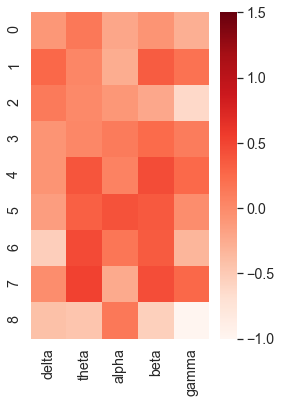}
         \caption{}
     \end{subfigure}
    \caption{Heatmap weights representations of the first (a) and the second (b) filter of a static convolutional layer at the end of the learning process on the SEED dataset.
    Filters are represented by matrices $W \in \mathbb{R}^{9 \times 5}$, where 9 is the kernel size and 5 is the number of input features, corresponding to the five EEG bands (see text for further details).}
    \label{fig:seed_viz_filter_static}
\end{figure}

\begin{figure}[!h]
    \centering
    \begin{subfigure}[b]{0.9\textwidth}
        \centering
        \includegraphics[width=0.31\linewidth]{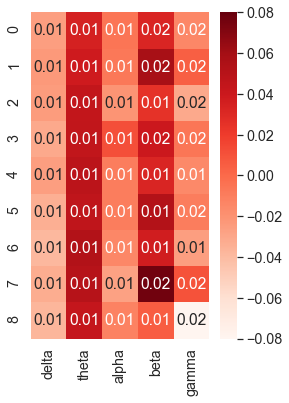}
        \hfill
        \includegraphics[width=0.31\linewidth]{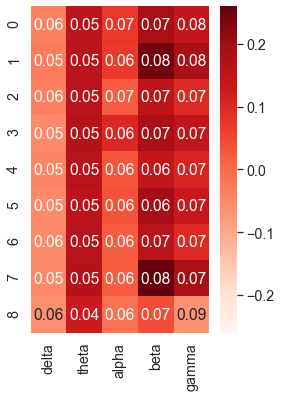}
        \hfill
        \includegraphics[width=0.31\linewidth]{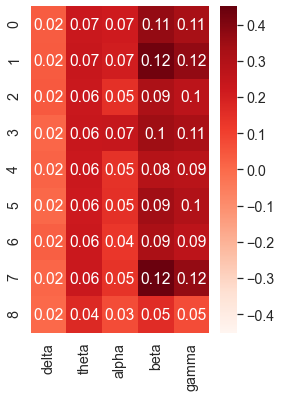}
        \caption{}
    \end{subfigure}
    \vskip\baselineskip
    \begin{subfigure}[b]{0.9\textwidth}
        \centering
        \includegraphics[width=0.31\linewidth]{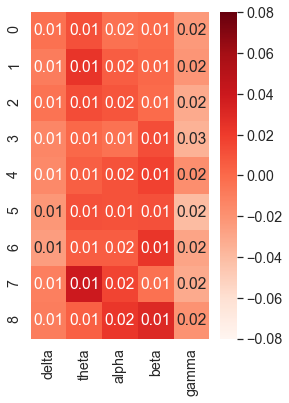}
        \hfill
        \includegraphics[width=0.31\linewidth]{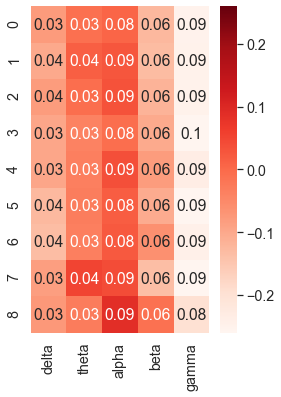}
        \hfill
        \includegraphics[width=0.31\linewidth]{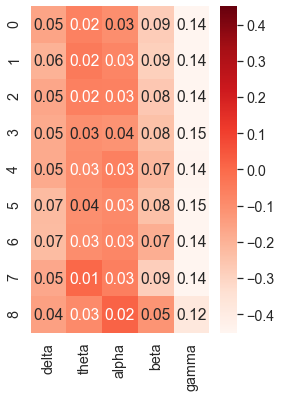}
        \caption{}
    \end{subfigure}
    \caption{Heatmap weights representations of the first (a) and the second (b) filter of a dynamic convolutional layer at the end of the learning process on the SEED dataset, averaged over the correctly classified samples. For each filter, averages related to the negative (left), neutral (center) and positive (right) labels are shown.
    Filters are represented by matrices $W \in \mathbb{R}^{9 \times 5}$, where 9 is the kernel size and 5 is the number of input features, corresponding to the five EEG bands (see text for further details).}
    \label{fig:seed_viz_filter_dynamic}
\end{figure}

A graphical representation of the filters related to the final configuration of each model was made using heatmaps in Figure \ref{fig:seed_viz_filter_static}, for the static model, and Figure \ref{fig:seed_viz_filter_dynamic}, for the dynamic one.
Considered a generic filter $W \in \mathbb{R}^{K \times J}$, the entry $W_{ij}$ represents the numerical value related to the transform of the $j$-th feature of the $i$-th neighbour. For the dynamic case, standard deviations of the weights are represented into the heatmap cells since, as we described above, for this analysis filters were averaged over a selected group of samples. 

In Figure \ref{fig:seed_viz_filter_dynamic} we can clearly confirm what we observed in Figure \ref{fig:seed_viz_filter_learning_dynamic}: the filter-generating network generates different filters according to the sample label.
It is interesting to notice how different features are enhanced in the label related filters: for example, for the positive label, both of the first and the second filter enhance, with high values in absolute value, features 3 and 4 (corresponding to beta and gamma bands), for each neighbor; differently, instead, filters related to the negative label enhances feature 1, corresponding to the theta band.
\section{Conclusion}
\label{sec:conclusion}
In this work, we have proposed a dynamic method to perform spatial convolution on graph-structured data. Combining the idea of having dynamically changeable behaviours in ANNs and convolutional graph neural networks, this work aimed to present a graph convolutional layer capable of performing convolution using node-specific filters in order to achieve an input-specific filtering operation. 
We have proposed a dynamic method to perform spatial convolution on graph-structured data in this work. Combining the idea of having dynamically changeable behaviours in ANNs and convolutional graph neural networks, this work aimed to present a graph convolutional layer capable of performing convolution using node-specific filters to achieve an input-specific filtering operation.
We altered the behaviour of a convolutional layer in a dynamic fashion using a filter-generating network. 
In this way, the proposed graph convolutional layer learns and applies input-specific filters, customising the filtering operation according to its input graph.

We run a series of experiments to assess the improvements in using a dynamic approach to generate convolutional filters.
It empirically emerged that our proposed strategy leads to better performances than those achieved using the static convolution on graphs. As we observed from Figures \ref{fig:seed_viz_filter_learning_dynamic} and \ref{fig:seed_viz_filter_dynamic}, the filter-generating networks learn to produce  class-specific filters, making the convolution operation input-specific actually.

Furthermore, it also emerged that convergence is reached in fewer epochs, reducing training time in a significant matter. Finally, using regularisation techniques, the use of an external module leads to filters having smaller weights than the static filters, which leads to an overall lower complexity of the main architecture.
These aspects were evident in the emotion recognition from EEG signals, which is a complex task to achieve.
In future work, we would like to introduce and analyse the use of a dynamic local filtering layer having local filters generated for each neighbourhood, as proposed by \citep{jia2016dynamic} in the image domain.

Currently, since the filter-generating network takes as input the entire graph, our strategy is constrained to data having a fixed graph topology. Using a local dynamic convolution, we could overcome this limit by extending this layer's functionality to data with non-fixed graphs topologies.

\bibliographystyle{elsarticle-harv} 
\bibliography{references}
\end{document}